\documentclass[11pt]{article}

\usepackage{spconf,amsmath,epsfig}
\usepackage{amsfonts}
\usepackage{amssymb}  
\usepackage[ruled,linesnumbered]{algorithm2e} 

\usepackage{float}

\usepackage{booktabs}
\usepackage[numbers,sort&compress]{natbib}
 
\usepackage{enumitem}
\setenumerate[1]{itemsep=0pt,partopsep=0pt,parsep=\parskip,topsep=0pt}
\setitemize[1]{itemsep=0pt,partopsep=0pt,parsep=\parskip,topsep=0pt}
\setdescription{itemsep=0pt,partopsep=0pt,parsep=\parskip,topsep=0pt}

\usepackage{multirow}
\usepackage{threeparttable}

\usepackage{graphicx}
\graphicspath{{images/}}    

\title{Semantic decoupled representation learning for remote sensing image change detection}

\name{Hao Chen\textsuperscript{1}, Yifan Zao\textsuperscript{1}, Liqin Liu\textsuperscript{1}, Song Chen\textsuperscript{2}, Zhenwei Shi\textsuperscript{1}  \thanks{Thanks to the National Natural Science Foundation of China under the Grants 62125102. \emph{Corresponding author: Zhenwei Shi (e-mail: shizhenwei@buaa.edu.cn).}}}
\address{\textsuperscript{1}Image Processing Center, School of Astronautics, Beihang University, Beijing 100191, China\\
\textsuperscript{2} Department of Journalism and Communications, Jeonuk National University, Jeonju-si 54896, South Korea\\
}
\begin{document}
%
\maketitle
\begin{abstract}
Contemporary transfer learning-based methods to alleviate the data insufficiency in change detection (CD) are mainly based on ImageNet pre-training. 
Self-supervised learning (SSL) has recently been introduced to remote sensing (RS) for learning in-domain representations.
Here, we propose a semantic decoupled representation learning for RS image CD. 
Typically, the object of interest (e.g., building) is relatively small compared to the vast background. Different from existing methods expressing an image into one representation vector that may be dominated by irrelevant land-covers, we disentangle representations of different semantic regions by leveraging the semantic mask. We additionally force the model to distinguish different semantic representations, which benefits the recognition of objects of interest in the downstream CD task. We construct a dataset of bitemporal images with semantic masks in an effortless manner for pre-training. Experiments on two CD datasets show our model outperforms ImageNet pre-training, in-domain supervised pre-training, and several recent SSL methods.

\end{abstract}
\section{Introduction}
\label{sec:intro}

Remote sensing (RS) image Change detection (CD) is the process of identifying changes of interest in RS images in the same geospatial region taken at different times. The key of CD is to identify real changes (e.g., buildings) while ignoring irrelevant changes (e.g., illumination, seasonal difference, and unconcerned land-cover changes).

Despite the great success of deep learning (DL)-based CD methods \cite{Shi2020}, the lack of a large labeled CD dataset limits their generalization on real-world applications. The most common way to handle data insufficiency is to fine-tune the model from ImageNet pre-training.
Considering the domain gap between natural and RS images, a new trend is to pre-train on the RS data to learn in-domain representations \cite{Zhang2020a,Neumann2020, Manas2021, Leenstra2021}. Supervised pre-training depends on labeled samples \cite{Zhang2020a, Neumann2020}, while self-supervised learning (SSL) makes use of unlabeled data \cite{Manas2021, Leenstra2021} by performing instance discrimination.

Most existing contrastive SSL methods \cite{Manas2021, Leenstra2021} express an image into one representation vector, which is usually coupled with objects of interest and backgrounds. Different from the object-centered natural image (i.e., ImageNet), small objects (e.g., buildings) typically present in various positions in an RS image. Simply performing contrastive learning on the coupled representation is not optimal because the background may dominate the representation due to class imbalance and make that of the object of interest ineffective.

We propose Semantic Decoupled Representation Learning (SDRL) to disentangle representations of the interest objects and others by leveraging the semantic mask to spatially pool the per-pixel image embeddings into several semantic vectors. We have two views for each semantic vector by applying synthetic image augmentations. We follow SimSiam \cite{Chen2020j} to implement the cross-view (i.e., a positive pair) similarity. We further utilize semantic relations to learn discriminative features by pushing away different semantic vectors. We collect in an effortless manner a dataset containing bi-temporal RS images and building masks for pre-training.

\begin{figure*}
        \centering
        \includegraphics[width=1\textwidth]{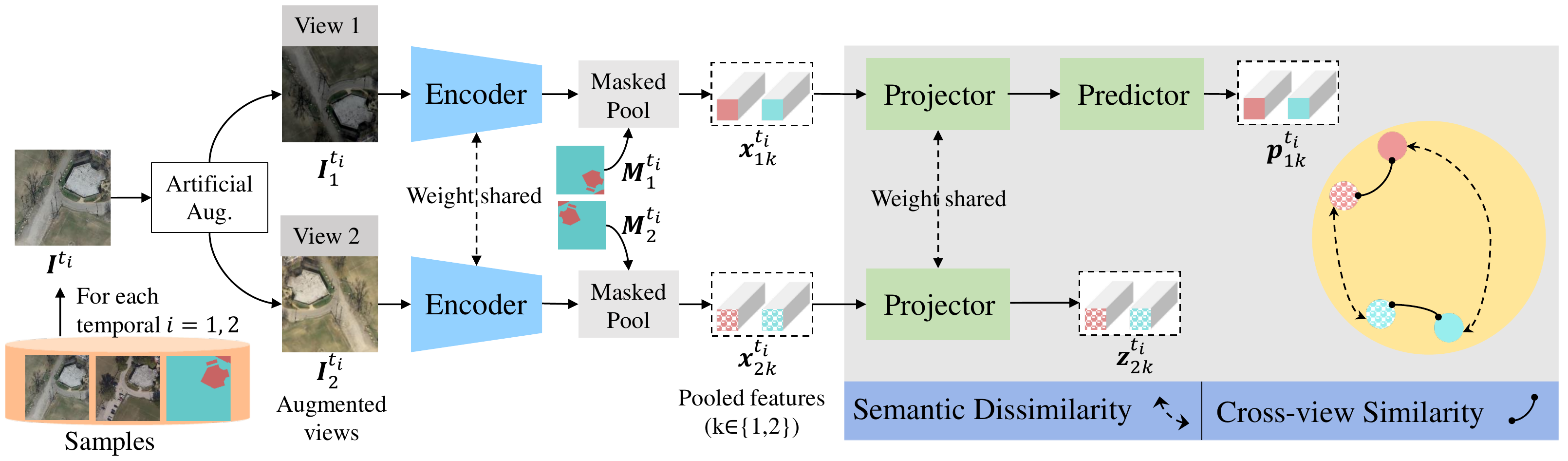}   
        \caption{Illustration of our semantic decoupled representation learning.}
        \label{fig:overall}
\end{figure*}

Our main contribution is to incorporate semantic masks into contrastive SSL to learn transferable features for the downstream CD. Experiments on two CD datasets \cite{Chen2020e, Ji2019a} demonstrate our method outperforms ImageNet, in-domain supervised pre-training, and several SSL methods, including the very recent SeCo \cite{Manas2021}.

\section{The Proposed Method}
\label{sec:method}

\subsection{Network Architecture}
Fig. \ref{fig:overall} illustrates the proposed Semantic Decoupled Representation Learning. Our architecture consists of a Siamese encoder/projector and a predictor. The Siamese encoder converts the input image views to semantic disentangled representations by leveraging semantic masks. Such intermediate representations are then projected into an embedding space by the projector. The predictor is employed to predict the representation of a view from that of another view.

\textbf{Views generation.}
Given an input sample consisting of a bitemporal image $\mathbf{I}^{t_{i}}, i\in \{1,2\}$ and a semantic mask $\mathbf{M}$, we randomly generate two views $\mathbf{I}_{j}^{t_{i}}, j\in \{1,2\}$ from each temporal image via synthetic augmentations. The semantic mask $\mathbf{M}_{j}^{t_{i}}$ is also generated by applying the same geometric augmentation as that in $\mathbf{I}_{j}^{t_{i}}$. 

\textbf{Disentangling semantic representations.} 
Two views $\mathbf{I}_{1}^{t_{i}},\mathbf{I}_{2}^{t_{i}}$ are then encoded by a Siamese encoder $f$ into dense representations $\mathbf{X}_{1}^{t_{i}}, \mathbf{X}_{2}^{t_{i}} \in \mathbb{R}^{H \times W \times C}$. Instead of pooling the per-pixel features into one vector, we decouple the global representation into several representation vectors by separately pooling features in each semantic region.
Formally, the disentangled representations $\mathbf{x}_{jk}^{t_{i}} \in \mathbb{R}^{C}, k\in \{1,2,...,\text{K}\}$ are calculated by
\begin{equation}
\setlength{\abovedisplayskip}{0.5pt}
\setlength{\belowdisplayskip}{0.5pt}
    \mathbf{x}_{j1}^{t_{i}},...,\mathbf{x}_{jK}^{t_{i}} = \text{MaskedPool}(\mathbf{X}_{j}^{t_{i}}, \mathbf{M}_{j}^{t_{i}}),
\end{equation}
where $\text{K}$ denotes the number of semantic categories. Specially, we use the binary mask to disentangle representations of foreground objects (i.e., buildings) and background. Therefore, we set $\text{K}=2$.  Note that semantic masks are resized to the same shape as the feature map before performing masked pooling.

\textbf{Semantic dissimilarity.}
Although representations of different semantic regions are decoupled, the relationship between them is not utilized. We employ a simple cosine-similarity-based loss to push away representations of different semantic regions. Our intuition is to enlarge the angle between the two vectors (i.e., representations of the foreground and the background) in the embedding space by minimizing their cosine similarity. 
Given the two embeddings $\mathbf{x}_{j1}^{t_{i}},\mathbf{x}_{j2}^{t_{i}}$ of a view $j$ of a temporal image $i$, the semantic dissimilar (SE) loss $\mathcal{L}_{sd,j}^{t_{i}}$ is:
\begin{equation}
\setlength{\abovedisplayskip}{0.5pt}
\setlength{\belowdisplayskip}{0.5pt}
    \mathcal{L}_{sd,j}^{t_{i}}= \mathcal{D}(\mathbf{x}_{j1}^{t_{i}}, \mathbf{x}_{j2}^{t_{i}}) + 1,
\end{equation}
where $\mathcal{D}(\cdot,\cdot)$ denotes the cosine similarity. 1 is supplemented to ensure a non-negative value.

In our data configuration, the semantic mask is paired with $\mathbf{I}^{t_{1}}$, but may not perfectly match $\mathbf{I}^{t_{2}}$ due to the inherent real changes. Therefore, we only use views from $\mathbf{I}^{t_{1}}$ to calculate the SE loss for one sample: $\mathcal{L}_{sd} =\frac{1}{2} \sum_{j=1}^{2}\mathcal{L}_{sd,j}^{t_{1}}$.

\textbf{Cross-view similarity.}
We follow SimSiam \cite{Chen2020j} to implement the cross-view similarity. Each semantic representation $\mathbf{x}_{jk}^{t_{i}}$ is projected into a space $\mathcal{Z} \in \mathbb{R}^{C'}$ by a MLP head (projector) $g$ to projection vectors $\mathbf{z}_{jk}^{t_{i}}$. Another MLP head (predictor) $h: \mathcal{Z} \mapsto \mathcal{Z}$ transforms one view (e.g., $\mathbf{z}_{1k}^{t_{i}}$) to a prediction vector (e.g., $\mathbf{p}_{1k}^{t_{i}} \in \mathbb{R}^{C'}$) and matches it to the other view (e.g., $\mathbf{z}_{2k}^{t_{i}}$). We minimize the negative cosine similarity of cross-view representations. A symmetric similarity loss is defined as:
\begin{equation}
\setlength{\abovedisplayskip}{0.5pt}
\setlength{\belowdisplayskip}{0.5pt}
    \mathcal{L}_{s,k}^{t_{i}} = 1-\frac{1}{2}(\mathcal{D}(\mathbf{p}_{1k}^{t_{i}}, \text{sg}(\mathbf{z}_{2k}^{t_{i}}) +  \mathcal{D}(\mathbf{p}_{2k}^{t_{i}}, \text{sg}(\mathbf{z}_{1k}^{t_{i}}))),
\end{equation}
where sg$(\cdot)$ means a stop-gradient operation. Previous work \cite{Chen2020j} shows such operation is vital to prevent model collapse. The similarity loss for one sample is given by $\mathcal{L}_{s} =\frac{1}{4}\sum_{i=1}^{2}\sum_{k=1}^{2}\mathcal{L}_{s,k}^{t_{i}}$.

\textbf{Overall loss} for one sample is given by: $\mathcal{L} = \mathcal{L}_{sd}+\mathcal{L}_{s}$. The total loss is averaged over all samples in a batch.

\subsection{Implementation Details}
\textbf{Encoder.} We use the ResNet-18 \cite{He2016} without fully connected layers (fc) and global pooling as $f$. We simply add a bilinear interpolation layer with an upsampling factor of 4 behind the ResNet-18 to reduce the loss of spatial details. The dimension $C$ of output is 512.

\textbf{Projector/Predictor.} $g, h$ are 2-layer MLPs with an output dimension $C^{'}=1024$. Their hidden dimensions are 1024 and 256. BN and ReLU are added between fc.

\textbf{Data augmentation.} We use similar augmentations in SimSiam, including color jittering, Gaussian blurring, random flip. Note that we do not apply the random crop.

\textbf{Optimizer.} We use SGD with a base learning rate of 0.01 and a poly decay schedule. We set the weight decay to 0.0005 and the SGD momentum to 0.9. The default number of epochs is 20 and the batch size is 64. 

\section{Experimental Results}
\label{sec:experiment}

\subsection{Experimental Setup}

\textbf{Pre-training Dataset.} We leverage image-label pairs from the existing Inria building dataset \cite{Maggiori2017}, which provides 180 labeled RS images ($5000\times 5000$, 0.3 m/pixel). The data is labeled into the building and non-building classes. For each image ($t_{1}$) in Inria, we obtain a co-registered image as temporal augmentation ($t_{2}$) via Google Earth. In this way, we collect bitemporal images with semantic building masks. We cut images into patches of size $256\times 256$ with no overlap and remove patches not containing building regions. It results in more than 45k patch samples. We randomly split it into training (80\%) and validation sets (20\%). Note that due to building changes over time, the semantic mask may not perfectly match the image of $t_{2}$.

\textbf{CD Datasets}. We conduct experiments on LEVIR-CD \cite{Chen2020e} and WHU-CD \cite{Ji2019a} to evaluate the pre-trained model. We follow \cite{Chen2021c} to split each dataset into training, validation, and testing sets. Each image is cut into small patches of size $256\times 256$ with no overlap.

\textbf{CD networks.} We employ a simple yet effective change detection model \cite{Chen2021a}. Differently, we use a more light FPN-based decoder head \cite{Lin2017a} for the feature extractor. Please refer to \cite{Chen2021a} for more implementation details.

\textbf{Evaluation Metrics.} We use the F1-score of the change category as the evaluation indices. 

Our models are implemented on PyTorch and trained using a single NVIDIA RTX 3090 GPU. 

\subsection{Overall Comparison}
\label{ssec:comparison}
We compare with several baselines, including random initialization, ImageNet pre-training, in-domain supervised pretraining (Sup.), and two SSL methods.

In-domain Sup.: We employ an FCN-based semantic segmentation network with a ResNet-18 backbone and an FPN head \cite{Lin2017a}, supervised by the image-mask pairs from our pre-training dataset. The best model on the validation set is used for the downstream task.

SimSiam \cite{Chen2020j}: A state-of-the-art contrastive method that does not require negative samples and uses stop-gradient and predictor to prevent model collapse. 

SeCo \cite{Manas2021}: A MoCo-based \cite{Chen2020d} method that uses multi-temporal RS images as natural augmentations to learn seasonal invariant/variant representations.

For a fair comparison, we implement these SSL methods using their public codes with default hyperparameters on our pre-training dataset.

We set a variety of data conditions: 1\%, 5\%, 20\%, and 100\%, each of which represents a proportion of available labeled training data. From Tab. \ref{tab:CD_SOTAs}, we can observe that the proposed method consistently outperforms other pre-trained models on the two CD datasets, especially in the small data regimes. Interestingly, we can achieve comparable results with only 20\% training data than the baseline (random) using 100\% data. It indicates our SDRL can effectively alleviate data insufficiency. Our model also outperforms the strong ImageNet pre-training the recent SeCo. We can observe in-domain sup. shows relatively poorer results on the WHU-CD dataset than on the LEVIR-CD dataset. It is because subdomain diversity may be present in different RS datasets. The traditional supervised pre-training may overfit patterns in a certain domain and lack transferability to another. Our empirical results indicate that our SDRL can learn more transferable features than supervised pre-training.

\begin{table}
    \centering
    \caption{Comparisons of pre-trained models on the LEVIR-CD and WHU-CD test sets. F1-score is given.}
    \resizebox{0.5\textwidth}{!}{
    \begin{tabular}{c|c|c|c|c|c|c|c|c}
   \toprule
    \multicolumn{1}{c}{} &
    \multicolumn{4}{c|}{\textbf{LEVIR-CD}}  &  \multicolumn{4}{c}{\textbf{WHU-CD}}  \\
    \multicolumn{1}{c}{} &
    \multicolumn{1}{|c|}{1\%} & \multicolumn{1}{|c|}{5\%}  & \multicolumn{1}{|c|}{20\%}  &  \multicolumn{1}{c|}{100\%} & \multicolumn{1}{|c|}{1\%} & \multicolumn{1}{|c|}{5\%}  & \multicolumn{1}{c|}{20\%}  &  \multicolumn{1}{|c}{100\%} \\
    \midrule
    Random & 21.32  & 47.05  & 80.30 & 87.63
    & 39.25  & 66.14 & 71.35 & 84.49 \\
    ImageNet & 24.33 & 65.18 & 85.76 & 88.16 & 
    32.42  & 73.52  & 81.98 &88.55\\
    In-domain Sup. & 29.29  & 71.96  & 85.68 & 88.06  &
    50.55 & 70.03 & 71.56 & 83.25 \\
    SimSiam \cite{Chen2020j} & 22.44 & 57.73 & 82.24 & 88.00 &
    42.09 & 66.10 & 73.41 & 84.69\\
    SeCo \cite{Manas2021} & 40.78 & 71.23 & 84.02 & 88.27  &
    51.68  & 72.15 & 80.00  & 88.09 \\
    \midrule
    Ours & \textbf{46.62} & \textbf{77.48}  &  \textbf{87.18} & \textbf{88.61}  & 
    \textbf{64.14}  & \textbf{76.50}& \textbf{84.17}  & \textbf{89.41} \\
   \bottomrule
   
    \end{tabular}}
    
    \label{tab:CD_SOTAs}
\end{table}

\begin{figure}
\begin{minipage}[t]{0.45\linewidth}
\centering
\includegraphics[width=\textwidth]{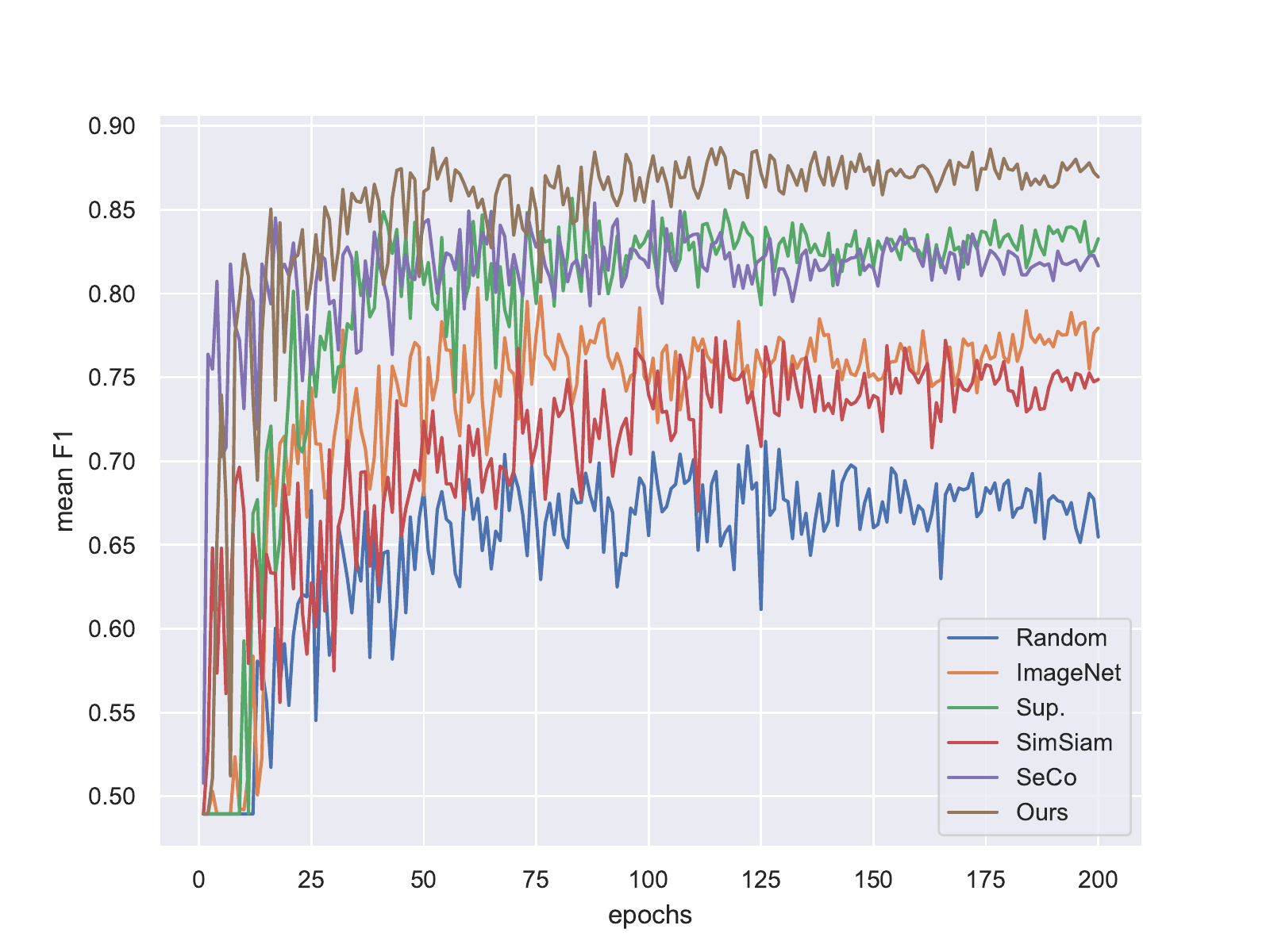}
  \centerline{(a) LEVIR-CD}

\end{minipage}
\begin{minipage}[t]{0.45\linewidth}
\centering
\includegraphics[width=\textwidth]{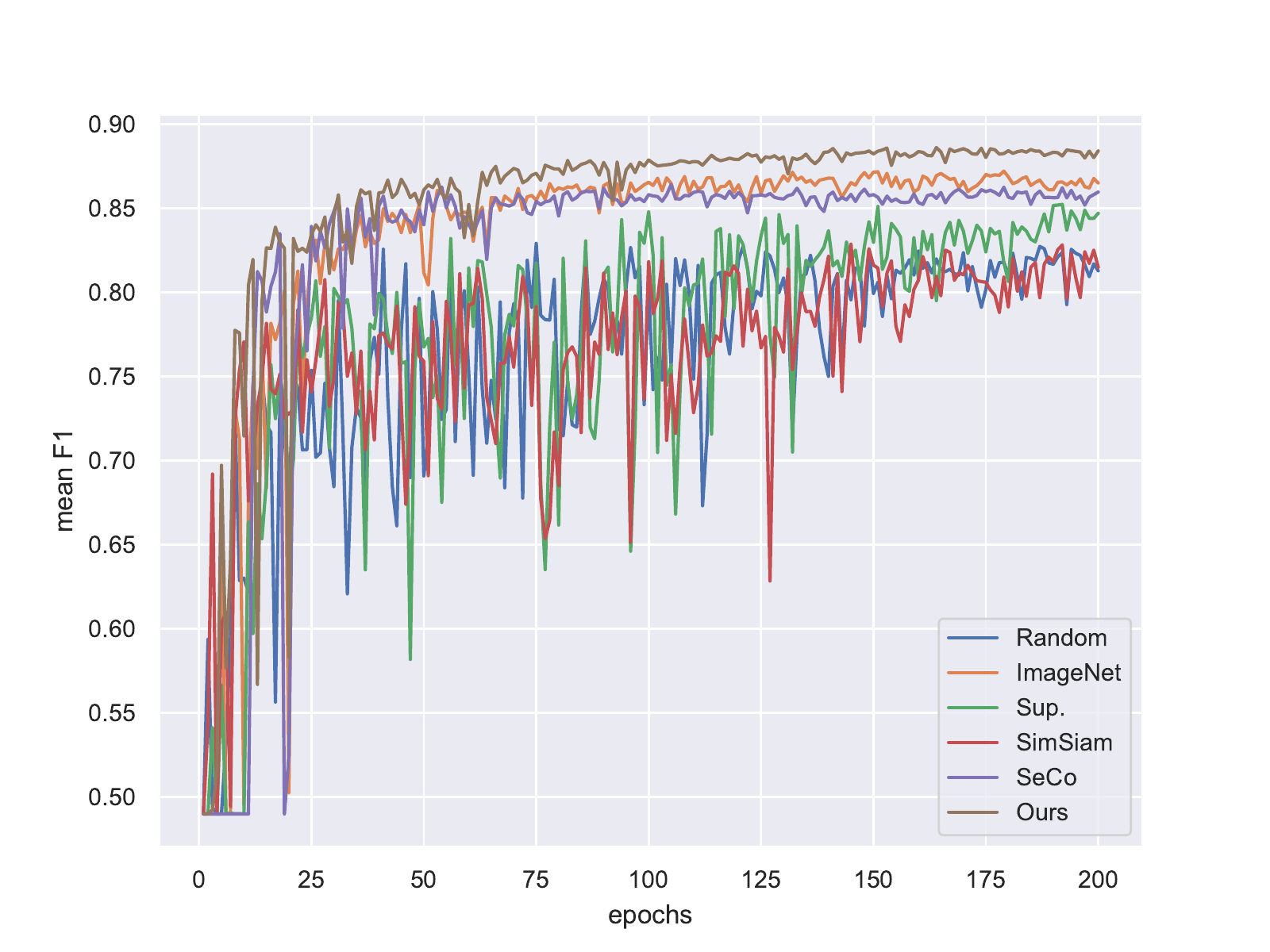}
  \centerline{(b) WHU-CD}

\end{minipage}
\caption{Accuracy of models for each training epoch. The mean F1-score is reported.}
\label{fig:training_acc}
\end{figure}

\textbf{Fast convergence.}
Fig. \ref{fig:training_acc} illustrates the validation accuracy (mean F1) for each epoch using 5\% LEVIR-CD/WHU-CD training data. Our SDRL outperforms others in terms of accuracy and stability. It indicates that SDRL accelerates fine-tuning convergence and incurs better performance on the downstream CD task.

\section{Conclusion}
\label{sec:conclusion}
We proposed a semantic decoupled pre-training method for RS image CD. We incorporate the semantic information into a contrastive learning framework to disentangle representations of different semantic regions (buildings and others). Semantic dissimilarity is utilized to guide the model to distinguish foreground from background. Experiments on two CD datasets verify the effectiveness of the proposed method. Our SDRL outperforms ImageNet, in-domain supervised pre-training, and several SSL pre-training methods. Empirical results indicate SDRL can well alleviate the data insufficiency in CD. We can achieve comparable results using only 20\% training data than baseline (random) using 100\% data.


{
\scriptsize
\bibliographystyle{IEEEbib}
\bibliography{refs}
}
\end{document}